\documentclass[11pt,a4paper]{article}
\usepackage[hyperref]{emnlp2018}
\usepackage{times}
\usepackage{latexsym}
\usepackage{graphicx}
\usepackage{bm}
\usepackage{multirow}

\usepackage{url}

\aclfinalcopy 

\title{LemmaTag: Jointly Tagging and Lemmatizing for Morphologically Rich Languages with BRNNs}

\author{
    Daniel Kondratyuk$^\dag$ \and
    Tom{\' a}{\v s} Gaven{\v c}iak$^\ddag$ \and
    Milan Straka$^\dag$ \and
    Jan Haji{\v c}$^\dag$ \\
    Charles University \\
    Faculty of Mathematics and Physics \\
    $^\dag$Institute of Formal and Applied Linguistics, $^\ddag$Department of Applied Mathematics \\
    {\tt\footnotesize dankondratyuk@gmail.com, gavento@kam.mff.cuni.cz, \{straka,hajic\}@ufal.mff.cuni.cz}
}

\date{}

\begin{document}

\maketitle

\begin{abstract}
We present LemmaTag, a featureless neural network architecture that jointly generates part-of-speech tags and lemmas for sentences by using bidirectional RNNs with character-level and word-level embeddings. We demonstrate that both tasks benefit from sharing the encoding part of the network, predicting tag subcategories, and using the tagger output as an input to the lemmatizer. We evaluate our model across several languages with complex morphology, which surpasses state-of-the-art accuracy in both part-of-speech tagging and lemmatization in Czech, German, and Arabic.
\end{abstract}

\section{Introduction}

Morphologically rich languages are often difficult to process in many NLP tasks \cite{tsarfaty2010statistical}. As opposed to analytical languages like English, morphologically rich languages encode diverse sets of grammatical information within each word using inflections, which convey characteristics such as case, gender, and tense. The addition of several inflectional variants across many words dramatically increases the vocabulary size, which results in data sparsity and out-of-vocabulary (OOV) issues.

\begin{figure}[b]
    \centering
    \includegraphics{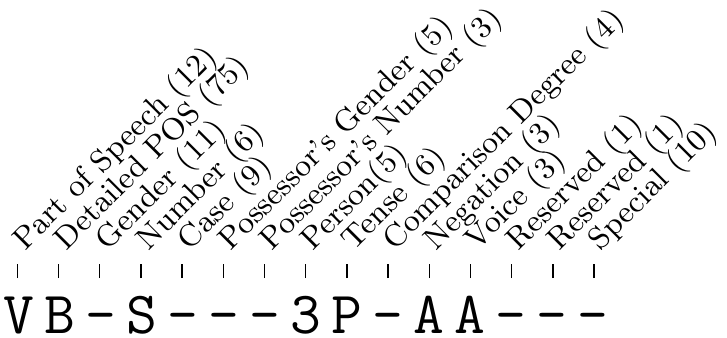}
    \caption{The tag components of the PDT Czech treebank with the numbers of valid values.
    Around 1500 different tags are in use in the PDT.}
    \label{fig:tag-components}
\end{figure}

Due to these issues, morphological part-of-speech (POS) tagging and lemmatization are heavily used in NLP tasks such as machine translation \cite{fraser2012modeling} and sentiment analysis \cite{abdul2014samar}. In morphologically rich languages, the POS tags typically consist of multiple morpho-syntactic subcategories providing additional information (see Figure~\ref{fig:tag-components}). Closely related to POS tagging is lemmatization, which involves transforming each word to its root or dictionary form. Both tasks require context-sensitive awareness to disambiguate words with the same form but different syntactic or semantic features and behavior. Furthermore, lemmatization of a word form can benefit substantially from the information present in morphological tags, as grammatical attributes often disambiguate word forms using context \cite{muller2015joint}.

We address context-sensitive POS tagging and lemmatization using a neural network model that jointly performs both tasks on each input word in a given sentence.\footnote{The code for this project is available at \url{https://github.com/hyperparticle/LemmaTag}} We train the model in a supervised fashion, requiring training data containing word forms, lemmas, and POS tags.  In addition, we incorporate the ideas from \newcite{inoue2017joint} to optionally allow the network to predict the subcategories of each tag to improve accuracy. Our model is related to the work of \newcite{muller2015joint}, which use conditional random fields (CRF) to jointly tag and lemmatize words for morphologically rich languages. 
The idea of jointly predicting several dimensions of categories has been explored prior to this work, for example, joint morphological and syntactic analysis \citep{bohnet:2013} or joint parsing and semantic role labeling \citep{gesmundo:2009}.

Our model consists of three parts: $(1)$ The \textbf{shared encoder}, which creates an internal representation for every word based on its character sequence and the sentence context. We adopt the encoder architecture of \newcite{chakrabarty2017context}, utilizing character-level \cite{heigold2017extensive} and word-level embeddings \cite{mikolov2013distributed, santos2014learning} processed through several layers of bidirectional recurrent neural networks (BRNN/BiRNN)~\cite{schuster1997bidirectional, chakrabarty2017context}. $(2)$ The \textbf{tagger decoder}, which applies a fully-connected layer to the outputs of the shared encoder to predict the POS tags. $(3)$ The \textbf{lemmatizer decoder}, which applies an RNN sequence decoder to the combined outputs of the shared encoder and tagger decoder, producing a sequence of characters that predict each lemma (similar to \newcite{bergmanis2018context}).

The main advantages over other proposed models are: $(i)$ The model is featureless, requiring little to no text preprocessing or morphological analysis postprocessing. $(ii)$ The model shares the word embeddings, character embeddings, and RNN encoder weights in the tagger and lemmatizer, improving both tagging and lemmatization accuracy while reducing the number of parameters required for both tasks. $(iii)$ The model predicts tag subcategories and provides the output of the tagger as features for the input of the lemmatizer, further improving accuracy.

We evaluate the accuracy of our model in POS tagging and lemmatization across several languages: Czech, Arabic, German, and English. For each language, we also compare the performance of a fully separate tagger and lemmatizer to the proposed joint model. Our results show that our joint model is able to improve the accuracy for both tasks, and achieves state-of-the-art performance in both POS tagging and lemmatization in Czech, German, and Arabic, while closely matching state-of-the-art performance for English.

\section{The Joint LemmaTag Model}

Given a sequence of words in a sentence $w_1, \dots, w_k$, the task of the model is to produce a sequence of associated tags $t_1, \dots, t_k$ and lemmas $\ell_1, \dots, \ell_k$. For a word $w_i$ at position $i$, we denote $c_{i,1}, c_{i,2} \dots c_{i,m_i}$ to be the sequence of characters that make up $w_i$, where $m_i$ indicates the length of the word string at position $i$. Analogously, we define $l_{i,1}, \dots l_{i,\lambda_i}$ to be the sequence of characters that make up the lemma $\ell_i$.

Our proposed model (shown in Figures~\ref{fig:embedding} and~\ref{fig:decoder}) is split into three parts: the shared encoder, the tagger, and the lemmatizer. The initial layers of the model are shared between the tagger and lemmatizer, encoding the words, characters, and context in a given sentence. The encoder then passes its outputs to two networks, which perform a classification task to predict tags by the tagger and a sequence prediction task to output lemmas (character-by-character) in the lemmatizer.

\subsection{Shared Encoder}

In the encoder shown in Figure~\ref{fig:embedding}, each character $c_{i,1}, c_{i,2} \dots c_{i,m_i}$ of a word $w_i$ is indexed into an embedding layer to produce fixed-length embedded vectors representing each character. These vectors are further passed into a layer of BRNNs composed of gated recurrent units (GRU) \cite{cho2014learning} producing outputs $\textbf{e}^c_{1}, \dots, \textbf{e}^c_{m}$, and whose final states are concatenated to produce the character-level embedding $\textbf{s}^c_i$ of the word. Similarly, we index $w_i$ into a word-level embedding layer to compute vector $\textbf{e}^b_i$. Then we sum these results to produce the final word embedding $\textbf{e}^w_i = \textbf{s}^c_i + \textbf{e}^b_i$.

We repeat this process independently for all the words in the sentence and feed the resulting sequence $\textbf{e}^w_1 \dots \textbf{e}^w_k$ into another two BRNN layers composed of long short-term memory units (LSTM) with residual connections. This produces word-level outputs $\textbf{o}^w_1, \dots \textbf{o}^w_k$ that encode sentence-level context for each word (we ignore the final hidden states). 

\subsection{Tagger}

\begin{figure}[t]
    \centering
    \includegraphics[scale=1.1]{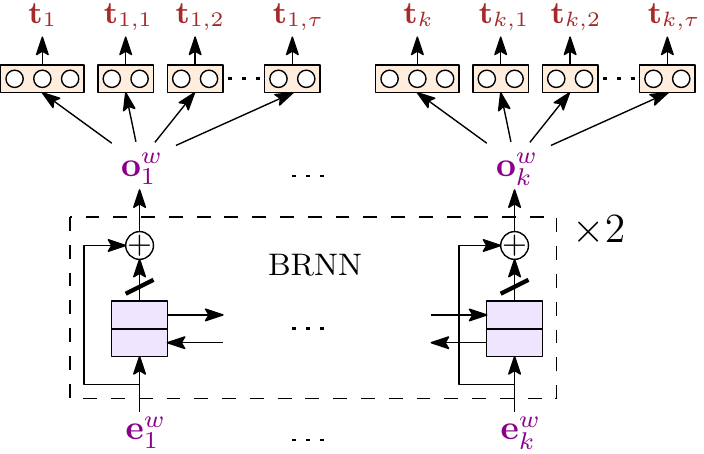}
    \vskip 8pt
    \hbox to \hsize{\cleaders\hbox to 1.8em{---}\hfill}
    \vskip 15pt
    \includegraphics[scale=1.1]{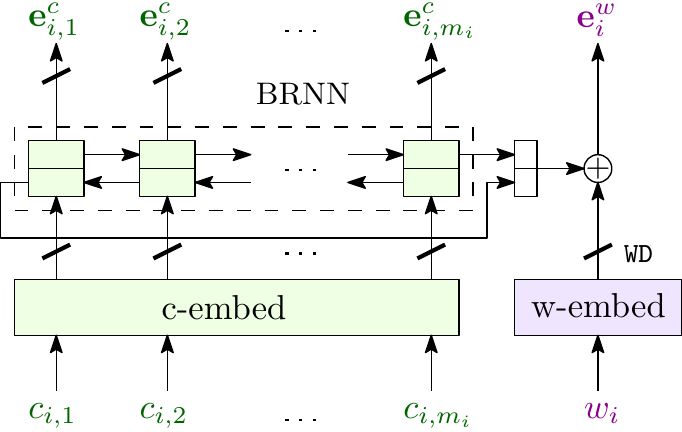}
    \caption{    \emph{Bottom:} Word-level encoder. The characters of every input word are embedded with a look-up table (c-embed) and encoded with a BRNN. The outputs $\textbf{e}^c_{i,j}$ are used in decoder attention, and the final states are summed with the word-level embedding (w-embed) to produce $\textbf{e}^w_i$. \texttt{WD} denotes word dropout.\\
    \emph{Top:} Sentence-level encoder and tag classifier. Two BRNN layers with residual connections act on the embedded words $\textbf{e}^w_i$ of a sentence, providing context. The output of the tag classification are the logits for both the whole tags $\textbf{t}_i$ and their components $\textbf{t}_{i,j}$.\\
    \emph{Both:} Thick slanted lines denote training dropout.}
    \label{fig:embedding}
\end{figure}

\begin{figure}[t]
    \centering
    \includegraphics[scale=1.2]{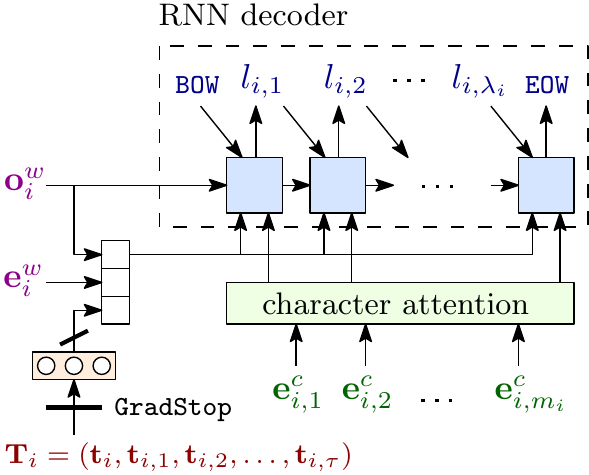}
    \caption{Lemma decoder, consisting of a standard seq2seq autoregressive decoder with Luong attention on character encodings, and with additional inputs of processed tagger features $\textbf{T}_i$, embeddings $\textbf{e}^w_i$ and sentence-level outputs $\textbf{o}^w_i$. Gradients from the lemmatizer are stopped from flowing into the tagger (denoted \texttt{GradStop}).}
    \label{fig:decoder}
\end{figure}

The task of the tagger is to predict a tag $t_i \in \mathcal{T}$ given a word $w_i$ and its context, where $\mathcal{T}$ is a set of possible tags. As explained the introduction, morphologically rich languages typically subdivide tags further into several subcategories $t_i = (t_{i,1}, \dots, t_{i,\tau})$, where $t_{i,j}\in\mathcal{T}_j$, the $j$-th subcategory. See Figure~\ref{fig:tag-components} for an illustration taken from the Czech PDT tagset where $\tau=15$.

Having the encoded words of a sentence available, the tagger consists of a fully-connected layer with $|\mathcal{T}|$ neurons whose input is the output of the word feature RNN $\textbf{o}^w_i$ (see figure~\ref{fig:embedding}). This layer produces the logits $\textbf{t}_i$ of the tag values and the predictions $t_i$ as the maximum-likelihood value (i.e., softmax).

To obtain the information about categorical nature of each tag, we also predict every category $t_{i,j}$ of the tag independently (if they exist in the dataset) with $\tau$ dense layers similar to \newcite{inoue2017joint}. The $j$-th layer has $|\mathcal{T}_j|$ neurons and outputs the logits $\textbf{t}_{i,j}$ for the category values. While these values are trained for, their value is not used in tag prediction. All tag values $\textbf{T}_i=(\textbf{t}_i, \textbf{t}_{i,1} \dots, \textbf{t}_{i,\tau})$ are concatenated into a flat vector and fed into the lemmatizer as an additional set of potentially useful features.

\subsection{Lemmatizer}

The task of the lemmatizer is to produce a sequence of characters $l_{i,1}, \dots, l_{i,\lambda_i}$ and the lemma length $\lambda_i$ for each lemma $\ell_i$. We use a recurrent sequence decoder, a setup typical of many sequence-to-sequence (seq2seq) tasks such as in neural machine translation \cite{sutskever2014sequence}.

The lemmatizer consists of a recurrent LSTM layer whose initial state is taken from word-level output $\textbf{o}^w_i$ and whose inputs consist of three parts. The first part is the embedding of the previous output character (initially a \emph{beginning-of-word} character \texttt{BOW}).

The second part is a character-level attention mechanism \cite{bahdanau2014neural} on the outputs of the character-level BRNN $\textbf{e}^c_{i,1}, \dots, \textbf{e}^c_{i,m_i}$. We employ the multiplicative attention mechanism described in \newcite{luong2015effective}, which allows the LSTM cell to compute an attention vector that selectively weights character-level information in $\textbf{e}^c_{i,j}$ at each time step $j$ based on the input state of the LSTM cell.

The third and final part of the RNN input allows the network to receive the information about the embedding of the word, the surrounding context of the sentence, and the output of the tagger. This output is the same for all time steps of a lemma and is a concatenation of the following: the output of the encoder $\textbf{o}^w_i$, the embedded word $\textbf{e}^w_i$ and processed tag features $\textbf{T}^f_i$. The tag features are obtained by projecting the concatenated outputs of the tagger $\textbf{T}_{i}$ through a fully connected layer with ReLU activation. During training, we do not pass the gradients back through $\textbf{T}_{i}$ to prevent the distortion of the tagger output.

The decoder performs greedy decoding to predict the character outputs. It runs until it produces the \emph{end-of-word} character \texttt{EOW} or reaches a character limit of $m_i+10$.

\begin{table*}[t!]
\begin{center}
\setlength{\tabcolsep}{4pt}
\begin{tabular}{|l|l|l|l|l|l|l|l|l|l|}
\hline \multirow{2}{*}{\bf Approach} & \multicolumn{2}{c|}{\bf Czech-PDT$^*$} & \multicolumn{2}{c|}{\bf \kern-.1emGerman-TIGER\kern-.1em} & \multicolumn{2}{c|}{\bf Arabic-PADT$^*$} & \multicolumn{2}{c|}{\bf Eng-EWT} & \bf \kern-.1em Eng-WSJ \kern-.3em\\ \cline{2-10}
& tag & lem & tag & lem & tag & lem & tag & lem & tag \\ \hline
LemmaTag (sep)   &     96.83 &     98.02 &     98.96 &     98.84 &     95.03 &     96.07 & \bf 95.50 &     97.03 &     97.59 \\
LemmaTag (joint) & \bf 96.90 & \bf 98.37 & \bf 98.97 & \bf 99.05 & \bf 95.21 & \bf 96.08 &     95.37 & \bf 97.53 &     N/A \\ \hline
\it SoTA results & \it 95.89$^a_+$ & \it 97.86$^b_+$ & \it 98.04$^c_+$ & \it 98.24$^c_+$ & \it 91.68$^d_+$ & \it 92.60$^e$ & \it 93.90$^e$ & \it 96.90$^e$ & \textit{\textbf{97.78}$^f_+$} \\
\hline
\end{tabular}
\end{center}
\caption{\label{table:final-results} Final accuracies on the test sets comparing the LemmaTag architecture as described (joint), LemmaTag neither sharing the encoder nor providing tagger features (sep), and the state-of-the-art results (SoTA). The state-of-the-art results are taken from the following papers: \textit{(a)}~\newcite{hajivc2009semi}, \textit{(b)}~\newcite{strakova14}, \textit{(c)}~\newcite{eger2016lemmatization}, \textit{(d)}~\newcite{inoue2017joint}, \textit{(e)} \newcite{straka2016udpipe}, \textit{(f)} \newcite{ling2015finding}. The results marked with a plus$_+$ use additional resources apart from the dataset, and datasets marked with a star$^*$ indicate the availability of subcategorical tags.}
\end{table*}

\subsection{Loss Function}


We define the final loss function as the weighted sum of the losses of the tagger and the lemmatizer:
$$L(\hat{y}, y) = \boldsymbol{\alpha} L(\hat{\textbf{y}}^{t}, \textbf{y}^{t}) + \beta L(\hat{\textbf{y}}^\ell, \textbf{y}^\ell)$$
where $\textbf{y}$ are the predicted outputs, $\hat{\textbf{y}}$ the expected outputs, $\textbf{y}^t$, the tag components and $\textbf{y}^\ell$ are the lemma characters. The tagger and lemmatizer losses are separately computed as the softmax cross entropy of the output logits. The weight hyperparameters $\boldsymbol{\alpha}, \beta$ scale the training losses so that the subtag and lemmatizer losses do not overpower the unfactored tag predictor gradients. The vector $\boldsymbol{\alpha}$ contains $\tau+1$ weights: one for the whole tag and one for every component.\footnote{If no components are available, $\tau=0$.}

\section{Experiments}

In this section, we show the outcomes of evaluation when running our joint tagger and lemmatizer and compare with the current state of the art in Czech, German, Arabic, and English datasets. Additionally, we evaluate the lemmatizer and tagger separately to compare the relative increase in tagging and lemmatization accuracy. 

\subsection{Datasets}

Our datasets consist of the Czech Prague Dependency Treebank (PDT) \cite{11858/00-097C-0000-0001-B098-5,11234/1-2621}, the German TIGER corpus \cite{brants2004tiger}, the Universal Dependencies Prague Arabic Dependency Treebank (UD-PADT) \cite{hajic2004prague}, the Universal Dependencies English Web Treebank (UD-EWT) \cite{silveira14gold}, and the WSJ portion of the English Penn Treebank (tags only) \cite{marcus1993building}. In all datasets, we use the tags specific to their respective language. Of these datasets, only Czech and Arabic provide subcategorical tags, and we use unfactored tags for the rest. See Table~\ref{table:final-results} for tagger and lemmatizer accuracies.

Note that the PDT dataset disambiguates lemmas with the same textual representation by appending a number as lemma sense indicator. For example, the dataset contains disambiguated lemmas \texttt{moc-1} (as \emph{power}) and \texttt{moc-2} (as \emph{too much}). About 17.5\% of the PDT tokens have such sense-disambiguated lemmas. LemmaTag predicts the lemmas including the senses and the accuracies in Table~\ref{table:final-results} take that into account. Ignoring the sense ambiguity, the lemmatization accuracy of the joint LemmaTag model is 98.94\% for Czech-PDT.

\subsection{Hyperparameters}

We use loss weights $\boldsymbol{\alpha}_0=1.0$ for the whole tags, $\boldsymbol{\alpha}_{1,\dots,\tau}=0.1$ for the tag component losses and $\beta = 0.5$ for the lemmatizer loss.\footnote{These are reasonable values to prevent gradients from overpowering one another. The lemmatizer tends to influence the tagger heavily.} The RNNs and word embedding tables have dimensionality 768 except for character-level embeddings and the character-level RNN, which are of dimension 384. The fully-connected layer whose inputs are $\textbf{T}_i$ is of dimension 256.

We train the models for 40 epochs with random permutations of training sentences and batches of 16 sentences. The starting learning rate is $\eta = 0.001$ and we scale this by 0.25 at epochs 20 and 30 to increase accuracy. We train the network using the lazy variant of the Adam optimizer \cite{kingma2014adam}, which only updates accumulators for variables that appear in the current batch \cite{tensorflow2018}, with parameters $\beta_1 = 0.9$ and $\beta_2 = 0.99$. We clip the global gradient norm to 3.0 to reduce the risk of exploding gradients.

To prevent the tagger from overfitting, we devise several strategies for regularization. We apply dropouts with rate 0.5 as indicated in Figures~\ref{fig:embedding} and \ref{fig:decoder}. The word dropout (\texttt{WD}) replaces 25\% of words by the unknown token \texttt{<unk>} to force the network to rely more on context, combatting data sparsity issues. Lastly, we employ label smoothing \cite{pereyra2017regularizing} which is a way to prevent the network from being too confident in any one class. The label smoothing parameter is set to $0.1$ for the tagger logits (both whole tags and the tag components).

Note that we did not perform any complex hyperparameter search. For additional information on real-world performance and additional techniques which have not improved evaluation accuracy, see Appendix~\ref{sec:supplemental}.

\section{Conclusion}

The evaluation results show that performing lemmatization and tagging jointly by sharing encoder parameters and utilizing tag features is mutually beneficial in morphologically rich languages. We have shown that incorporating these ideas results in excellent performance, surpassing state-of-the-art in Czech, German, and Arabic POS tagging and lemmatization by a substantial margin, while closely matching state-of-the-art English POS tagging accuracy.

However, in languages with weak morphology such as English (and German to a lesser extent), sharing the encoder parameters may even hurt the performance of the tagger. We believe this is a consequence of tags correlating less with word-level morphology, and more with sentence-level syntax in morphologically poor languages. Lemma prediction could benefit from the syntactic information in the tags, but the tag predictions rely more on syntactic structure (i.e., word order) rather than on root forms of individual words which could be ambiguous.

There are some possible performance improvements and additional metrics which we leave for future work. For simplicity, one improvement we intentionally left out is the use of additional data. We can incorporate word2vec \cite{mikolov2013efficient} or ELMo \cite{peters2018deep} word representations, which have shown to reduce out-of-domain issues and provide semantic information \cite{eger2016lemmatization}. A second improvement is to integrate information from a morphological dictionary to resolve certain ambiguities \cite{hajivc2009semi, inoue2017joint}. A third improvement can be to replace the seq2seq lemmatizer decoder with a classifier that chooses a corresponding edit tree to modify (reduce) the word form to its lemma \cite{chakrabarty2017context}. A fourth possible improvement would be to experiment with the Transformer model \cite{vaswani2017attention}, which utilizes non-recurrent multi-headed self-attention and has been shown to achieve state-of-the-art performance in several related sequence tasks \cite{dehghani2018universal}. Lastly, we would like to evaluate LemmaTag on a wider range of languages, e.g., on the Universal Dependencies \cite{nivreUD} languages and treebanks which employ lemmatization, and to analyze the use of different types of POS tags in the model.

The code we used for LemmaTag is available at \url{https://github.com/hyperparticle/LemmaTag}.

\section*{Acknowledgments}

The work described herein has been supported by the City of Prague under the ``OP PPR'' program, project No. CZ.07.1.02/0.0/0.0/16\_023/0000108 and it has been using language resources developed by the LINDAT/CLARIN project of the Ministry of Education, Youth and Sports of the Czech Republic (project LM2015071).

Tom{\' a}{\v s} Gaven{\v c}iak has been supported by Czech Science Foundation (GACR) project 17-10090Y ``Network optimization''.
Daniel Kondratyuk has been supported by the Erasmus Mundus program in Language \& Communication Technologies (LCT).

\bibliography{emnlp2018}

\begin{thebibliography}{36}
\expandafter\ifx\csname natexlab\endcsname\relax\def\natexlab#1{#1}\fi

\bibitem[{Abdul-Mageed et~al.(2014)Abdul-Mageed, Diab, and
  K{\"u}bler}]{abdul2014samar}
Muhammad Abdul-Mageed, Mona Diab, and Sandra K{\"u}bler. 2014.
\newblock Samar: Subjectivity and sentiment analysis for arabic social media.
\newblock \emph{Computer Speech \& Language}, 28(1):20--37.

\bibitem[{Bahdanau et~al.(2014)Bahdanau, Cho, and Bengio}]{bahdanau2014neural}
Dzmitry Bahdanau, Kyunghyun Cho, and Yoshua Bengio. 2014.
\newblock Neural machine translation by jointly learning to align and
  translate.
\newblock \emph{arXiv preprint arXiv:1409.0473}.

\bibitem[{Bergmanis and Goldwater(2018)}]{bergmanis2018context}
Toms Bergmanis and Sharon Goldwater. 2018.
\newblock Context sensitive neural lemmatization with lematus.
\newblock In \emph{Proceedings of the 2018 Conference of the North American
  Chapter of the Association for Computational Linguistics: Human Language
  Technologies, Volume 1 (Long Papers)}, volume~1, pages 1391--1400.

\bibitem[{Bohnet et~al.(2013)Bohnet, Nivre, Boguslavsky, Farkas, Ginter, and
  Hajič}]{bohnet:2013}
Bernd Bohnet, Joakim Nivre, Igor Boguslavsky, Richárd Farkas, Filip Ginter,
  and Ja~n Hajič. 2013.
\newblock Joint morphological and syntactic analysis for richly inflected
  languages.
\newblock \emph{Transactions of the Association for Computational Linguistics},
  1:415--428.

\bibitem[{Brants et~al.(2004)Brants, Dipper, Eisenberg, Hansen-Schirra,
  K{\"o}nig, Lezius, Rohrer, Smith, and Uszkoreit}]{brants2004tiger}
Sabine Brants, Stefanie Dipper, Peter Eisenberg, Silvia Hansen-Schirra, Esther
  K{\"o}nig, Wolfgang Lezius, Christian Rohrer, George Smith, and Hans
  Uszkoreit. 2004.
\newblock Tiger: Linguistic interpretation of a german corpus.
\newblock \emph{Research on language and computation}, 2(4):597--620.

\bibitem[{Chakrabarty et~al.(2017)Chakrabarty, Pandit, and
  Garain}]{chakrabarty2017context}
Abhisek Chakrabarty, Onkar~Arun Pandit, and Utpal Garain. 2017.
\newblock Context sensitive lemmatization using two successive bidirectional
  gated recurrent networks.
\newblock In \emph{Proceedings of the 55th Annual Meeting of the Association
  for Computational Linguistics (Volume 1: Long Papers)}, volume~1, pages
  1481--1491.

\bibitem[{Cho et~al.(2014)Cho, Van~Merri{\"e}nboer, Gulcehre, Bahdanau,
  Bougares, Schwenk, and Bengio}]{cho2014learning}
Kyunghyun Cho, Bart Van~Merri{\"e}nboer, Caglar Gulcehre, Dzmitry Bahdanau,
  Fethi Bougares, Holger Schwenk, and Yoshua Bengio. 2014.
\newblock Learning phrase representations using rnn encoder-decoder for
  statistical machine translation.
\newblock \emph{arXiv preprint arXiv:1406.1078}.

\bibitem[{Dehghani et~al.(2018)Dehghani, Gouws, Vinyals, Uszkoreit, and
  Kaiser}]{dehghani2018universal}
Mostafa Dehghani, Stephan Gouws, Oriol Vinyals, Jakob Uszkoreit, and {\L}ukasz
  Kaiser. 2018.
\newblock Universal transformers.
\newblock \emph{arXiv preprint arXiv:1807.03819}.

\bibitem[{Eger et~al.(2016)Eger, Gleim, and Mehler}]{eger2016lemmatization}
Steffen Eger, R{\"u}diger Gleim, and Alexander Mehler. 2016.
\newblock Lemmatization and morphological tagging in german and latin: A
  comparison and a survey of the state-of-the-art.
\newblock In \emph{LREC}.

\bibitem[{Fraser et~al.(2012)Fraser, Weller, Cahill, and
  Cap}]{fraser2012modeling}
Alexander Fraser, Marion Weller, Aoife Cahill, and Fabienne Cap. 2012.
\newblock Modeling inflection and word-formation in smt.
\newblock In \emph{Proceedings of the 13th Conference of the European Chapter
  of the Association for Computational Linguistics}, pages 664--674.
  Association for Computational Linguistics.

\bibitem[{Gesmundo et~al.(2009)Gesmundo, Henderson, Merlo, and
  Titov}]{gesmundo:2009}
Andrea Gesmundo, James Henderson, Paola Merlo, and Ivan Titov. 2009.
\newblock A latent variable model of synchronous syntactic-semantic parsing for
  multiple languages.
\newblock In \emph{Proceedings of the Thirteenth Conference on Computational
  Natural Language Learning: Shared Task}, CoNLL '09, pages 37--42,
  Stroudsburg, PA, USA. Association for Computational Linguistics.

\bibitem[{Haji{\v c} et~al.(2018)Haji{\v c}, Bej{\v c}ek, B{\'e}mov{\'a},
  Bur{\'a}{\v n}ov{\'a}, Haji{\v c}ov{\'a}, Havelka, Homola, K{\'a}rn{\'{\i}}k,
  Kettnerov{\'a}, Klyueva, Kol{\'a}{\v r}ov{\'a}, Ku{\v c}ov{\'a},
  Lopatkov{\'a}, Mikulov{\'a}, M{\'{\i}}rovsk{\'y}, Nedoluzhko, Pajas,
  Panevov{\'a}, Pol{\'a}kov{\'a}, Rysov{\'a}, Sgall, Spoustov{\'a}, Stra{\v
  n}{\'a}k, Synkov{\'a}, {\v S}ev{\v c}{\'{\i}}kov{\'a}, {\v S}t{\v
  e}p{\'a}nek, Ure{\v s}ov{\'a}, Vidov{\'a}~Hladk{\'a}, Zeman,
  Zik{\'a}nov{\'a}, and {\v Z}abokrtsk{\'y}}]{11234/1-2621}
Jan Haji{\v c}, Eduard Bej{\v c}ek, Alevtina B{\'e}mov{\'a}, Eva Bur{\'a}{\v
  n}ov{\'a}, Eva Haji{\v c}ov{\'a}, Ji{\v r}{\'{\i}} Havelka, Petr Homola,
  Ji{\v r}{\'{\i}} K{\'a}rn{\'{\i}}k, V{\'a}clava Kettnerov{\'a}, Natalia
  Klyueva, Veronika Kol{\'a}{\v r}ov{\'a}, Lucie Ku{\v c}ov{\'a}, Mark{\'e}ta
  Lopatkov{\'a}, Marie Mikulov{\'a}, Ji{\v r}{\'{\i}} M{\'{\i}}rovsk{\'y}, Anna
  Nedoluzhko, Petr Pajas, Jarmila Panevov{\'a}, Lucie Pol{\'a}kov{\'a},
  Magdal{\'e}na Rysov{\'a}, Petr Sgall, Johanka Spoustov{\'a}, Pavel Stra{\v
  n}{\'a}k, Pavl{\'{\i}}na Synkov{\'a}, Magda {\v S}ev{\v c}{\'{\i}}kov{\'a},
  Jan {\v S}t{\v e}p{\'a}nek, Zde{\v n}ka Ure{\v s}ov{\'a}, Barbora
  Vidov{\'a}~Hladk{\'a}, Daniel Zeman, {\v S}{\'a}rka Zik{\'a}nov{\'a}, and
  Zden{\v e}k {\v Z}abokrtsk{\'y}. 2018.
\newblock Prague dependency treebank 3.5.
\newblock {LINDAT}/{CLARIN} digital library at the Institute of Formal and
  Applied Linguistics ({{\'U}FAL}, \url{http://hdl.handle.net/11234/1-2621}),
  Faculty of Mathematics and Physics, Charles University.

\bibitem[{Haji{\v c} et~al.(2006)Haji{\v c}, Panevov{\'a}, Haji{\v c}ov{\'a},
  Sgall, Pajas, {\v S}t{\v e}p{\'a}nek, Havelka, Mikulov{\'a}, {\v
  Z}abokrtsk{\'y}, {\v S}ev{\v c}{\'{\i}}kov{\'a}-Raz{\'{\i}}mov{\'a}, and
  Ure{\v s}ov{\'a}}]{11858/00-097C-0000-0001-B098-5}
Jan Haji{\v c}, Jarmila Panevov{\'a}, Eva Haji{\v c}ov{\'a}, Petr Sgall, Petr
  Pajas, Jan {\v S}t{\v e}p{\'a}nek, Ji{\v r}{\'{\i}} Havelka, Marie
  Mikulov{\'a}, Zden{\v e}k {\v Z}abokrtsk{\'y}, Magda {\v S}ev{\v
  c}{\'{\i}}kov{\'a}-Raz{\'{\i}}mov{\'a}, and Zde{\v n}ka Ure{\v s}ov{\'a}.
  2006.
\newblock Prague {D}ependency {T}reebank 2.0 ({PDT} 2.0).
\newblock {LINDAT}/{CLARIN} digital library at the Institute of Formal and
  Applied Linguistics ({{\'U}FAL}), Faculty of Mathematics and Physics, Charles
  University.

\bibitem[{Haji{\v{c}} et~al.(2009)Haji{\v{c}}, Raab, Spousta
  et~al.}]{hajivc2009semi}
Jan Haji{\v{c}}, Jan Raab, Miroslav Spousta, et~al. 2009.
\newblock Semi-supervised training for the averaged perceptron pos tagger.
\newblock In \emph{Proceedings of the 12th Conference of the European Chapter
  of the Association for Computational Linguistics}, pages 763--771.
  Association for Computational Linguistics.

\bibitem[{Hajic et~al.(2004)Hajic, Smrz, Zem{\'a}nek, {\v{S}}naidauf, and
  Be{\v{s}}ka}]{hajic2004prague}
Jan Hajic, Otakar Smrz, Petr Zem{\'a}nek, Jan {\v{S}}naidauf, and Emanuel
  Be{\v{s}}ka. 2004.
\newblock Prague arabic dependency treebank: Development in data and tools.
\newblock In \emph{Proc. of the NEMLAR Intern. Conf. on Arabic Language
  Resources and Tools}, pages 110--117.

\bibitem[{Heigold et~al.(2017)Heigold, Neumann, and van
  Genabith}]{heigold2017extensive}
Georg Heigold, Guenter Neumann, and Josef van Genabith. 2017.
\newblock An extensive empirical evaluation of character-based morphological
  tagging for 14 languages.
\newblock In \emph{Proceedings of the 15th Conference of the European Chapter
  of the Association for Computational Linguistics: Volume 1, Long Papers},
  volume~1, pages 505--513.

\bibitem[{Inoue et~al.(2017)Inoue, Shindo, and Matsumoto}]{inoue2017joint}
Go~Inoue, Hiroyuki Shindo, and Yuji Matsumoto. 2017.
\newblock Joint prediction of morphosyntactic categories for fine-grained
  arabic part-of-speech tagging exploiting tag dictionary information.
\newblock In \emph{Proceedings of the 21st Conference on Computational Natural
  Language Learning (CoNLL 2017)}, pages 421--431.

\bibitem[{Kingma and Ba(2014)}]{kingma2014adam}
Diederik~P Kingma and Jimmy Ba. 2014.
\newblock Adam: A method for stochastic optimization.
\newblock \emph{arXiv preprint arXiv:1412.6980}.

\bibitem[{Ling et~al.(2015)Ling, Lu{\'\i}s, Marujo, Astudillo, Amir, Dyer,
  Black, and Trancoso}]{ling2015finding}
Wang Ling, Tiago Lu{\'\i}s, Lu{\'\i}s Marujo, Ram{\'o}n~Fernandez Astudillo,
  Silvio Amir, Chris Dyer, Alan~W Black, and Isabel Trancoso. 2015.
\newblock Finding function in form: Compositional character models for open
  vocabulary word representation.
\newblock \emph{arXiv preprint arXiv:1508.02096}.

\bibitem[{Luong et~al.(2015)Luong, Pham, and Manning}]{luong2015effective}
Minh-Thang Luong, Hieu Pham, and Christopher~D Manning. 2015.
\newblock Effective approaches to attention-based neural machine translation.
\newblock \emph{arXiv preprint arXiv:1508.04025}.

\bibitem[{Marcus et~al.(1993)Marcus, Marcinkiewicz, and
  Santorini}]{marcus1993building}
Mitchell~P Marcus, Mary~Ann Marcinkiewicz, and Beatrice Santorini. 1993.
\newblock Building a large annotated corpus of english: The penn treebank.
\newblock \emph{Computational linguistics}, 19(2):313--330.

\bibitem[{Mikolov et~al.(2013{\natexlab{a}})Mikolov, Chen, Corrado, and
  Dean}]{mikolov2013efficient}
Tomas Mikolov, Kai Chen, Greg Corrado, and Jeffrey Dean. 2013{\natexlab{a}}.
\newblock Efficient estimation of word representations in vector space.
\newblock \emph{arXiv preprint arXiv:1301.3781}.

\bibitem[{Mikolov et~al.(2013{\natexlab{b}})Mikolov, Sutskever, Chen, Corrado,
  and Dean}]{mikolov2013distributed}
Tomas Mikolov, Ilya Sutskever, Kai Chen, Greg~S Corrado, and Jeff Dean.
  2013{\natexlab{b}}.
\newblock Distributed representations of words and phrases and their
  compositionality.
\newblock In \emph{Advances in neural information processing systems}, pages
  3111--3119.

\bibitem[{M{\"u}ller et~al.(2015)M{\"u}ller, Cotterell, Fraser, and
  Sch{\"u}tze}]{muller2015joint}
Thomas M{\"u}ller, Ryan Cotterell, Alexander Fraser, and Hinrich Sch{\"u}tze.
  2015.
\newblock Joint lemmatization and morphological tagging with lemming.
\newblock In \emph{Proceedings of the 2015 Conference on Empirical Methods in
  Natural Language Processing}, pages 2268--2274.

\bibitem[{Nivre et~al.(2016)Nivre, de~Marneffe, Ginter, Goldberg, Hajic,
  Manning, McDonald, Petrov, Pyysalo, Silveira, Tsarfaty, and Zeman}]{nivreUD}
Joakim Nivre, Marie-Catherine de~Marneffe, Filip Ginter, Yoav Goldberg, Jan
  Hajic, Christopher~D. Manning, Ryan McDonald, Slav Petrov, Sampo Pyysalo,
  Natalia Silveira, Reut Tsarfaty, and Daniel Zeman. 2016.
\newblock Universal dependencies v1: A multilingual treebank collection.
\newblock In \emph{Proceedings of the Tenth International Conference on
  Language Resources and Evaluation (LREC 2016)}, Paris, France. European
  Language Resources Association (ELRA).

\bibitem[{Pereyra et~al.(2017)Pereyra, Tucker, Chorowski, Kaiser, and
  Hinton}]{pereyra2017regularizing}
Gabriel Pereyra, George Tucker, Jan Chorowski, {\L}ukasz Kaiser, and Geoffrey
  Hinton. 2017.
\newblock Regularizing neural networks by penalizing confident output
  distributions.
\newblock \emph{arXiv preprint arXiv:1701.06548}.

\bibitem[{Peters et~al.(2018)Peters, Neumann, Iyyer, Gardner, Clark, Lee, and
  Zettlemoyer}]{peters2018deep}
Matthew~E Peters, Mark Neumann, Mohit Iyyer, Matt Gardner, Christopher Clark,
  Kenton Lee, and Luke Zettlemoyer. 2018.
\newblock Deep contextualized word representations.
\newblock \emph{arXiv preprint arXiv:1802.05365}.

\bibitem[{Santos and Zadrozny(2014)}]{santos2014learning}
Cicero~D Santos and Bianca Zadrozny. 2014.
\newblock Learning character-level representations for part-of-speech tagging.
\newblock In \emph{Proceedings of the 31st International Conference on Machine
  Learning (ICML-14)}, pages 1818--1826.

\bibitem[{Schuster and Paliwal(1997)}]{schuster1997bidirectional}
Mike Schuster and Kuldip~K Paliwal. 1997.
\newblock Bidirectional recurrent neural networks.
\newblock \emph{IEEE Transactions on Signal Processing}, 45(11):2673--2681.

\bibitem[{Silveira et~al.(2014)Silveira, Dozat, de~Marneffe, Bowman, Connor,
  Bauer, and Manning}]{silveira14gold}
Natalia Silveira, Timothy Dozat, Marie-Catherine de~Marneffe, Samuel Bowman,
  Miriam Connor, John Bauer, and Christopher~D. Manning. 2014.
\newblock A gold standard dependency corpus for {E}nglish.
\newblock In \emph{Proceedings of the Ninth International Conference on
  Language Resources and Evaluation (LREC-2014)}.

\bibitem[{Straka et~al.(2016)Straka, Hajic, and Straková}]{straka2016udpipe}
Milan Straka, Jan Hajic, and Jana Straková. 2016.
\newblock Udpipe: Trainable pipeline for processing conll-u files performing
  tokenization, morphological analysis, pos tagging and parsing.
\newblock In \emph{Proceedings of the Tenth International Conference on
  Language Resources and Evaluation (LREC 2016)}, Paris, France. European
  Language Resources Association (ELRA).

\bibitem[{Strakov\'{a} et~al.(2014)Strakov\'{a}, Straka, and
  Haji\v{c}}]{strakova14}
Jana Strakov\'{a}, Milan Straka, and Jan Haji\v{c}. 2014.
\newblock Open-{S}ource {T}ools for {M}orphology, {L}emmatization, {POS}
  {T}agging and {N}amed {E}ntity {R}ecognition.
\newblock In \emph{Proceedings of 52nd Annual Meeting of the Association for
  Computational Linguistics: System Demonstrations}, pages 13--18, Baltimore,
  Maryland. Association for Computational Linguistics.

\bibitem[{Sutskever et~al.(2014)Sutskever, Vinyals, and
  Le}]{sutskever2014sequence}
Ilya Sutskever, Oriol Vinyals, and Quoc~V Le. 2014.
\newblock Sequence to sequence learning with neural networks.
\newblock In \emph{Advances in neural information processing systems}, pages
  3104--3112.

\bibitem[{TensorFlow(2018)}]{tensorflow2018}
TensorFlow. 2018.
\newblock tf.contrib.opt.lazyadamoptimizer: Class lazyadamoptimizer.
\newblock TensorFlow documentation from tensorflow.org.

\bibitem[{Tsarfaty et~al.(2010)Tsarfaty, Seddah, Goldberg, K{\"u}bler, Candito,
  Foster, Versley, Rehbein, and Tounsi}]{tsarfaty2010statistical}
Reut Tsarfaty, Djam{\'e} Seddah, Yoav Goldberg, Sandra K{\"u}bler, Marie
  Candito, Jennifer Foster, Yannick Versley, Ines Rehbein, and Lamia Tounsi.
  2010.
\newblock Statistical parsing of morphologically rich languages (spmrl): what,
  how and whither.
\newblock In \emph{Proceedings of the NAACL HLT 2010 First Workshop on
  Statistical Parsing of Morphologically-Rich Languages}, pages 1--12.
  Association for Computational Linguistics.

\bibitem[{Vaswani et~al.(2017)Vaswani, Shazeer, Parmar, Uszkoreit, Jones,
  Gomez, Kaiser, and Polosukhin}]{vaswani2017attention}
Ashish Vaswani, Noam Shazeer, Niki Parmar, Jakob Uszkoreit, Llion Jones,
  Aidan~N Gomez, {\L}ukasz Kaiser, and Illia Polosukhin. 2017.
\newblock Attention is all you need.
\newblock In \emph{Advances in Neural Information Processing Systems}, pages
  5998--6008.

\end{thebibliography}
\bibliographystyle{acl_natbib_nourl}

\appendix

\section{Appendix}
\label{sec:supplemental}

\subsection{GPU Performance}

We ran all the tests on an NVIDIA GTX 1080 Ti GPU. The joint LemmaTag training takes about 3 hours for Arabic PADT, 4.5 hours for English EWT, 12 hours for German TIGER, and 22 hours for Czech PDT. The separate models take about 50\% more time. After training, the lemma and tag predictions of 219,000 test tokens of the Czech PDT take about 100 seconds.


\subsection{Other Techniques}

We briefly summarize some of the additional techniques we have tried but which do not improve the results. While some of those techniques do help on smaller models or earlier in the training, the effect on the fully trained network seems to be marginal or even detrimental.

\emph{Separate sense prediction.} Instead of predicting the sense disambiguation with the lemmatizer (Czech only), we tried to predict sense as an additional classification problem with one dense layer based on $\textbf{o}^w_{i}$ and $\textbf{T}_i$, but it seems to perform slightly worse (0.2\%). 

\emph{Beam search decoder.} We have implemented a beam search decoder for the lemmatizer instead of the standard greedy one, but the improvement was marginal (around 0.01\%).

\emph{Variational dropout.} While the dropouts in the LemmaTag are completely random, variational dropout erases the same channels across the time steps of the RNN. While this generally improves training in convolutional networks and RNNs, we saw no significant difference.

\emph{Layer normalization.} Layer normalization applied to the encoding RNNs did not bring significant gain and also slowed down the training.

\end{document}